\documentclass[acmtog]{acmart} 

\acmSubmissionID{1022}

\usepackage{booktabs} 
\usepackage{cleveref}

\usepackage[ruled]{algorithm2e} 

\SetAlFnt{\small}
\SetAlCapFnt{\small}
\SetAlCapNameFnt{\small}
\SetAlCapHSkip{0pt}

\copyrightyear{2024}
\acmYear{2024}
\setcopyright{rightsretained}
\acmConference[SA Conference Papers '24]{SIGGRAPH Asia 2024 Conference Papers}{December 3--6, 2024}{Tokyo, Japan}
\acmBooktitle{SIGGRAPH Asia 2024 Conference Papers (SA Conference Papers '24), December 3--6, 2024, Tokyo,
Japan}\acmDOI{10.1145/3680528.3687681}
\acmISBN{979-8-4007-1131-2/24/12}

\usepackage{xcolor}         
\definecolor{coltoncolor}{RGB}{255, 87, 51}

\definecolor{mkcolor}{RGB}{255,0, 128}

\definecolor{adamcolor}{RGB}{0,128,0}

\definecolor{leocolor}{RGB}{0,0,255}

\begin{document}
\title{Dynamic Gaussian Marbles for Novel View Synthesis of Casual Monocular Videos}

\author{Colton Stearns}
\orcid{0000-0002-3297-2870}
\affiliation{%
 \institution{Stanford University}
 \city{Stanford}
 \state{CA}
 \postcode{94305}
 \country{USA}}
\email{coltongs@stanford.edu}
\author{Adam W. Harley}
\orcid{0000-0002-9851-4645}
\affiliation{%
 \institution{Stanford University}
  \city{Stanford}
 \state{CA}
 \postcode{94305}
 \country{USA}
}
\email{aharley@cs.stanford.edu}
\author{Mikaela Uy}
\orcid{0009-0009-4917-7724}
\affiliation{
\institution{Stanford University}
 \city{Stanford}
 \state{CA}
 \postcode{94305}
 \country{USA}
}
\email{mikacuy@stanford.edu}
\author{Florian Dubost}
\orcid{0000-0002-7035-2680}
\affiliation{
\institution{Google}
 \city{Mountain View}
 \country{USA}
}
\email{fdubost@google.com}
\author{Federico Tombari}
\orcid{0000-0001-5598-5212}
\affiliation{%
 \institution{Google}
 \city{Zurich}
 \country{Switzerland}
}
\email{tombari@google.com}
\author{Gordon Wetzstein}
\orcid{0000-0002-9243-6885}
\affiliation{%
 \institution{Stanford University}
  \city{Stanford}
 \state{CA}
 \postcode{94305}
 \country{USA}
}
\email{gordon.wetzstein@stanford.edu}
\author{Leonidas Guibas}
\orcid{0000-0002-8315-4886}
\affiliation{%
 \institution{Stanford University}
  \city{Stanford}
 \state{CA}
 \postcode{94305}
 \country{USA}
}

\begin{teaserfigure}
  \includegraphics[width=\textwidth]{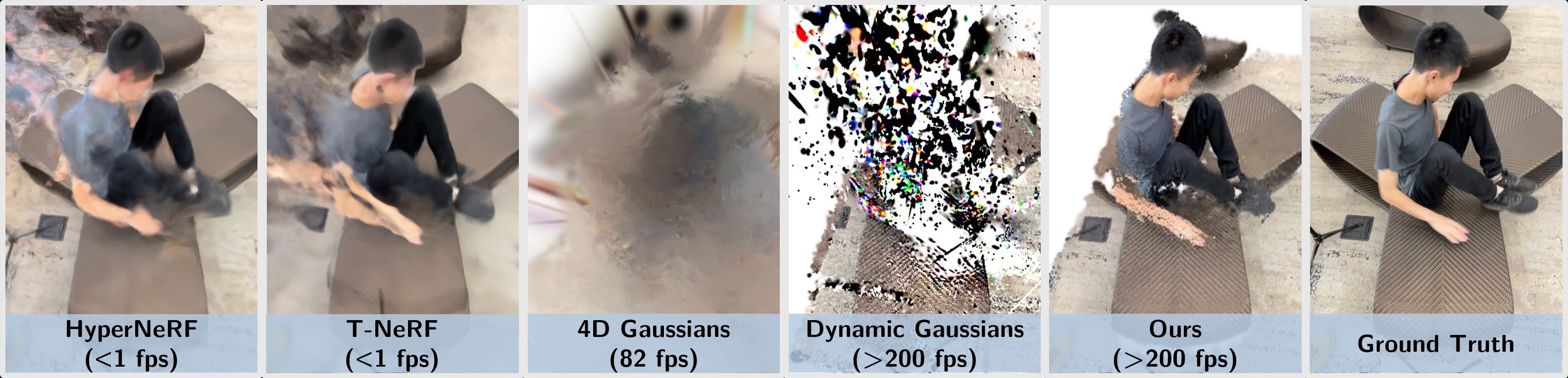}
  \caption{Our method achieves high-quality novel-view synthesis given a challenging monocular video as input. In contrast, other Gaussian representations arrive at poor local minima, 
  while NeRF methods are on-par but exhibit slow rendering, poor tracking, and lack an editable and compositional structure.}
  \label{fig:teaser}
\end{teaserfigure}

\renewcommand\shortauthors{Stearns, C. et al}

\begin{abstract}

Gaussian splatting has become a popular representation for novel-view synthesis, exhibiting clear strengths in efficiency, photometric quality, and compositional edibility. 
Following its success, many works have extended Gaussians to 4D, showing that dynamic Gaussians maintain these benefits while also tracking scene geometry far better than alternative representations.  
Yet, 
these methods assume dense multi-view videos as supervision, constraining their use 
to controlled capture settings. 
In this work, we are interested in extending the capability of Gaussian scene representations to casually captured monocular videos. 
We show that existing 4D Gaussian methods dramatically fail in this setup because the monocular setting is underconstrained.
%
%
Building off this finding, we propose a method we call \textit{Dynamic Gaussian Marbles}, which consist of three core modifications that target the difficulties of the monocular setting.
First, we use isotropic Gaussian ``marbles'', reducing the degrees of freedom of each Gaussian, and constraining the optimization to focus on motion and appearance over local shape.
Second, we employ a hierarchical divide-and-conquer learning strategy to efficiently guide the optimization towards solutions with globally coherent motion. 
Finally, we add image-level and geometry-level priors into the optimization, including a tracking loss that takes advantage of recent progress in point tracking.
By constraining the optimization in these ways, Dynamic Gaussian Marbles learns Gaussian trajectories that enable novel-view rendering and accurately capture the 3D motion of the scene elements.
We evaluate on the (monocular) Nvidia Dynamic Scenes dataset and the Dycheck iPhone dataset, 
and show that Gaussian Marbles significantly outperforms other Gaussian baselines in quality, and is on-par with non-Gaussian representations, all 
while maintaining the efficiency, compositionality,  editability, and tracking benefits of Gaussians.

\end{abstract}

%
%



\maketitle

\section{Introduction}
It is very challenging to convert everyday monocular videos of dynamic scenes into reconstructions which are renderable from alternative viewpoints. 
Doing so seems to require extracting 3D geometry, motion, and radiance, all from pixels alone.  
Achieving this in a robust way would greatly extend current capabilities in video production, 3D content creation, virtual reality, and synthetic data generation, as well as advance computer vision. 

In recent years, the research community has made tremendous progress in building renderable 3D representations from \textit{multi-view captures}. In particular, Gaussian Splatting~\cite{kerbl20233Dgaussians} has emerged as a leading solution for novel-view synthesis of \textit{static} scenes. By representing the 3D space with a collection of 3D Gaussians and ``splatting'' these onto the image plane, Gaussian Splatting achieves high-quality photometric reconstruction and efficient rendering. Furthermore, the Gaussian representation exhibits desirable compositional capabilities: a scene can be edited by, for example, moving (or removing) the Gaussians that make up an object. 
%

Many works have since extended Gaussian Splatting to the 4D setting, allowing \textit{dynamic} scenes to be reconstructed in a manner that 3D content is tracked and rendered with impressive accuracy~\cite{luiten2023dynamic, huang2023sparsecontrolled, Duisterhof2023MDSplattingLM, lin2023gaussianflow}.  
However, these works are designed for the setting where there are \textit{multiple simultaneous} viewpoints of a scene (i.e., a multi-camera setup), which limits their use to purpose-built capture environments. 

In this work, we are interested in using Gaussians for simple, casual, \textit{monocular} captures, where a single camera is being moved smoothly about a dynamic scene (e.g., by a human). 
%
Our core finding is that while current methods for dynamic or 4D Gaussians are highly underconstrained in the absence of multi-view information, we can recover suitable constraints through a careful optimization strategy, off-the-shelf estimations of depth and motion, and geometry-based regularizations on scene structure. We demonstrate our findings in a method we call Dynamic Gaussian Marbles (or `Gaussian Marbles' for short).
%

Concretely, Gaussian Marbles introduces changes to the core representation, the learning strategy, and the objective function, all with the aim of guiding the optimization process to arrive at solutions that better generalize to novel views. 
First, Gaussian Marbles removes the anisotropic nature of typical Gaussians, and simply uses isotropic ``marbles'' -- we find that marbles are better suited for the underconstrained monocular setting.
%
%
%
Second, Gaussian Marbles uses a divide-and-conquer learning algorithm -- we divide a long video into subsequences and optimize each subsequence independently, and then merge pairs of subsequences.
This accounts for the fact that it is easier to solve for motion and geometry within shorter time horizons, and converts long-sequence tracking into a task of gluing together neighboring subsequences. 
%
Third, Gaussian Marbles uses freely-available priors in both image space and 3D space. In the image plane, we use off-the-shelf models SegmentAnything ~\cite{kirillov2023segany, yang2023trackanything}, CoTracker ~\cite{karaev2023cotracker}, and DepthAnything ~\cite{yang2024depthanything},
and guide our 3D representation according to these 2D cues. 
In 3D space, we regularize Gaussian trajectories with rigidity and Chamfer priors. 

%
We show that Gaussian Marbles greatly outperforms other dynamic Gaussian methods in the casual monocular setting. We evaluate on the Nvidia Dynamic Scenes dataset, the DyCheck iPhone dataset, and the Total-Recon dataset. 
Furthermore, we show that we are on-par with NeRF-based methods, while retaining key advantages of  
efficient rendering, tracking, and editability.

\begin{figure*}
    \centering
    \includegraphics[width=\textwidth]{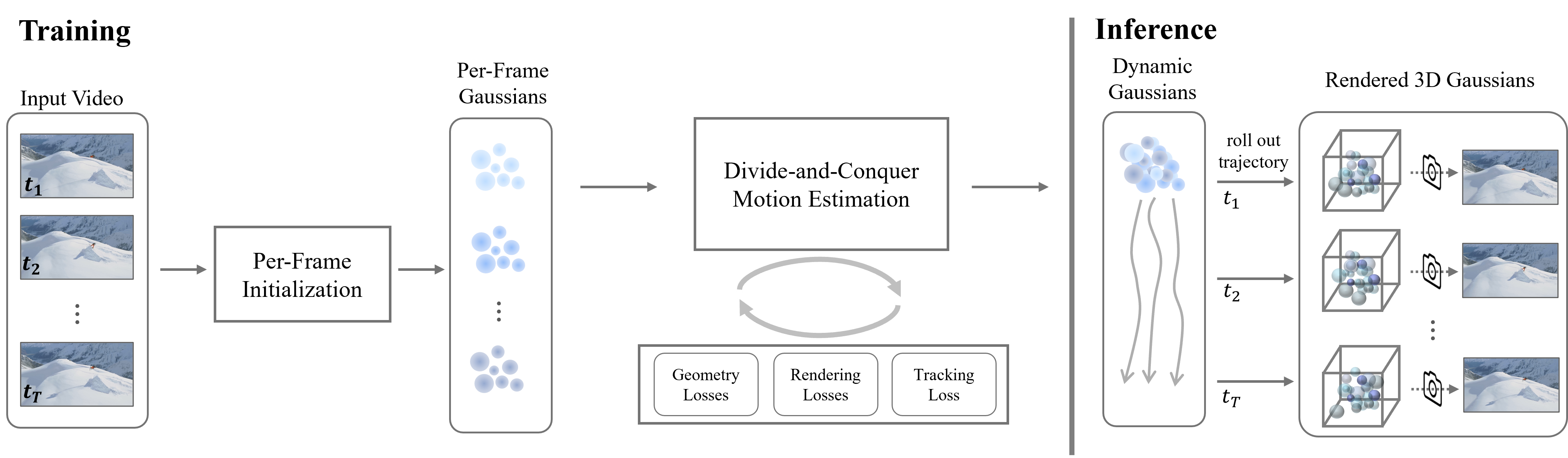}
    \caption{\textbf{Gaussian Marbles Overview.}
    At training time (left), we take as input a video and optimize a Gaussian-based reconstruction of the data. We begin by initializing a set of Gaussians for each frame. Then, we employ a bottom-up divide-and-conquer strategy to merge sets of Gaussians, which iteratively attributes longer motion trajectories to Gaussians. Motion estimation is achieved by optimizing a rendering loss (i.e., color reconstruction), a tracking loss (i.e., Gaussians should move similarly to point tracks), and geometry-based losses (e.g., local rigidity). After training (right), each Gaussian has a multi-frame trajectory,  
    and we can render into a timestep using the set of Gaussian trajectories that span the timestep.
    }
    \label{fig:method-overview}
\end{figure*}
\section{Related Work}
\subsection{Gaussian Splatting}
Gaussian-based representations have a long history in modeling 3D scenes~\cite{blinn1982generalization} due to their efficiency, interpretability, and compositionality. 
The key idea is to represent a scene using a set of anisotropic Gaussians, each equipped with opacity and color attributes. The Gaussians enable efficient color rendering and are also well suited to applications such as scene editing and pose estimation~\cite{keselman2022approximate,keselman2023flexible}. 
In the past year, Gaussian scene representations have received renewed attention due to 3D Gaussian Splatting (3DGS)~\cite{kerbl20233Dgaussians}, which efficiently and differentiably``splats'' the Gaussians onto the image plane.
%
Many works have dived deeper on the advantages of 3DGS, including its compositionality~\cite{gaussian_grouping, yu2023cogs}, speed~\cite{lee2023compact, morgenstern2023compact, niedermayr2023compressed} and quality~\cite{lee2024deblurring, Zhang2024FreGS3G, Feng2024ANS}, and it has also been adapted into many downstream applications such as pose estimation~\cite{fan2024instantsplat}, SLAM~\cite{Matsuki:Murai:etal:CVPR2024}, semantic scene understanding~\cite{zhou2023feature, cen2023saga, gaussian_grouping}, human avatars~\cite{li2024animatablegaussians, qian20233dgsavatar}, text-to-3D~\cite{tang2023dreamgaussian, yi2023gaussiandreamer}, and more ~\cite{Dai2024GaussianSurfels}.

\subsection{Gaussians for Dynamic Scenes}
Many works have begun extending the 3DGS representation to the 4D domain, aiming to solve the challenge of dynamic scene reconstruction. 
One popular direction is to model the 4D content using a set of 3D Gaussian trajectories through time~\cite{luiten2023dynamic, sun20243dgstream, Duisterhof2023MDSplattingLM}, and to learn trajectories by sequentially optimizing for per-Gaussian offsets into the next frame. 
Some recent works propose a compact set of trajectories via sparse control points~\cite{huang2023sparsecontrolled} or an explicit motion basis~\cite{das2023neuralparametric, Katsumata2023AnE3, lin2023gaussianflow, li2023spacetimetrajs, yu2023cogs}.
These trajectory-based methods often exhibit impressive tracking of scene geometry.

Another line of works~\cite{wu20234dgaussians, yang2023deformable3dgs, Liang2023GauFReGD, Guo2024Motionaware3G} uses a a time-conditioned deformation network to warp a canonical set of Gaussians into each timeframe. While this is a more efficient and compact representation of motion, it can be particularly challenging to learn the appropriate deformations and to avoid local minima.
Lastly, a few works~\cite{Duan20244DGS, yang2023gs4d} directly model Gaussians that extend across space and time, i.e. Gaussians with mean and covariances in 4-D.

While Gaussian Marbles is more similar to works that represent motion as 3D Gaussian trajectories, we note that all these previous methods largely address the same multi-camera setting. In contrast, our work tackles the more challenging monocular setting. We note that there are concurrent works~\cite{das2023neuralparametric, Katsumata2023AnE3} that tackle the pseudo-monocular setting and showcase results on datasets with ``teleporting'' cameras or large amounts of effective multi-view information -- please refer to DyCheck~\cite{gao2022dycheck} for a thorough overview of this phenomena. In contrast, our approach is intended for \textit{any} casual monocular video.

\subsection{Other Neural Scene Representations for Dynamic Scenes}
Many earlier works explored neural scene representations for dynamic scenes. One family of works is the extension of neural radiance fields~\cite{mildenhall2020nerf} (NeRFs) in 4D~\cite{Ramasinghe2024blirf, Wang2021NeuralTF, song2023nerfplayer, li2020neuralsceneflow, Gao2021DynNeRF, gao2022dycheck, Cao2023HEXPLANE, Fridovich2023kplanes, bui2023dyblurf, Jang2022DTensoRFTR}, treating time as a fourth dimension and as an additional coordinate in the neural field.
Another approach is to combine a ``canonical'' 3D NeRF with a time-conditioned deformation field ~\cite{liu2023robust, fang2022TiNeuVox, park2021nerfies, park2021hypernerf, johnson2022ub4d, kirschstein2023nersemble, Wang2023flowsupervision, tretschk2020nonrigid}. The deformation field can help disentangle motion and geometry, resulting in a more constrained and better-behaved scene. 
A few works explored NeRF-based representations for the casual monocular setting -- DyNiBar and MonoNeRF~\cite{li2023dynibar, tian2023mononerf} showed compelling results by combining NeRF with image-based rendering~\cite{wang2021ibrnet}, and Wang et al.~\cite{Wang2024DiffusionPF} used diffusion to regularize a 4D NeRF~\cite{Cao2023HEXPLANE, Fridovich2023kplanes}.
Finally, some concurrent works explore alternate plane-based and feed-forward representations for the casual monocular setting  ~\cite{zhao2024pgdvs, lee2023casual-fvs}.

In contrast to these representations, dynamic Gaussians have advantages in tracking, fast rendering, and compositional editability.

\section{Preliminaries} \label{sec:prelims}

\subsection{3D Gaussian Splatting}
3D Gaussian splatting~\cite{kerbl20233Dgaussians} is a differentiable rendering pipeline that represents a scene as a collection of 3D Gaussians and ``splats'' them onto the image plane. Concretely, a 3D scene is represented by 3D Gaussians, $\mathcal{G}$, with each Gaussian parameterized by its mean $\mu \in \mathbb{R}^3$, rotation $R \in \mathbb{R}^{3\times3}$, scale $s \in \mathbb{R}^3$, color $c \in \mathbb{R}^3$, and opacity $\alpha \in \mathbb{R}$. The scale and rotation can be composed into a 3D covariance matrix, $\Sigma = RSS^TR^T$, where $S$ is the $3\times3$ diagonal scaling matrix.

Given 3D Gaussians $\mathcal{G}$ and a camera viewing transformation $W$, the covariance matrix in camera coordinates can be computed as $\Sigma' = J W \Sigma W^T J^T$, where $J$ is the Jacobian of the approximately-affine projective transformation. Given camera-aligned Gaussians, the pipeline executes an efficient differentiable tile-based rasterization -- the image is divided into tiles $16\times16$, and for each tile, the influencing Gaussians are sorted by depth and alpha-composited in the image plane. In contrast to volumetric rendering approaches~\cite{mildenhall2020nerf}, Gaussian Splatting is extremely efficient and often renders over $100$ times faster than its volumetric counterpart.

\section{Method}
\begin{figure}
    \centering
    \includegraphics[width=0.49\textwidth]{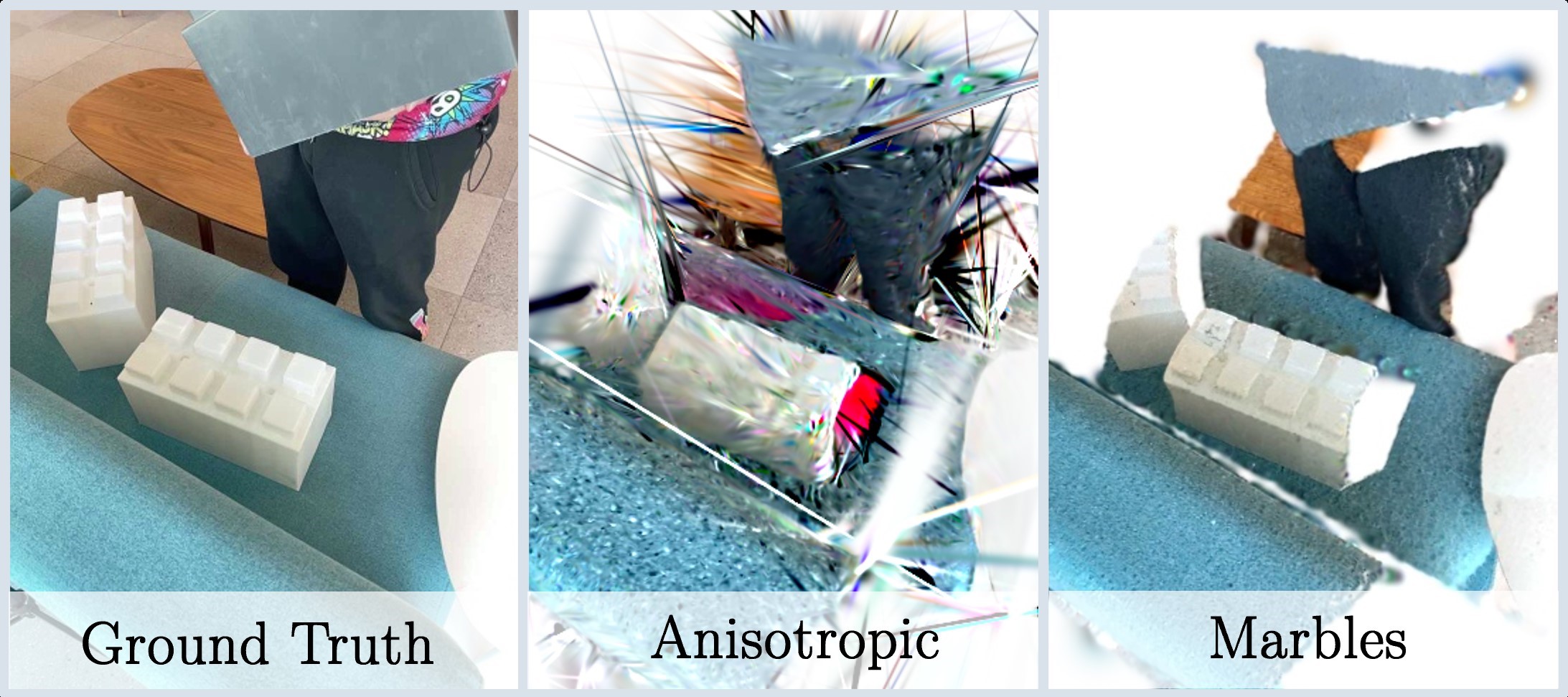}
    \caption{We train anisotropic Gaussians and our Gaussian Marbles for 100K iterations on a single monocular RGBD image. While the training view reconstruction is perfect for both, anistropic Gaussians lead to undesirable artifacts in novel views, whereas Gaussian marbles generalize well.}
    \label{fig:why-marbles}
\end{figure}

We provide an overview of Gaussian Marbles in ~\Cref{fig:method-overview}. We take as input a casually captured monocular video (i.e., a sequence of images captured by a single camera traversing a dynamic scene). We begin by initializing a set of Gaussian ``marbles'' for each frame, and we consider these initial marbles to have trajectories of length 1. Then, we seek to merge these disjoint sets of short-trajectory marbles into longer trajectories. 
To build up trajectories, we employ a bottom-up divide-and-conquer merging strategy as depicted in \Cref{fig:hierarchical-motion}: 
we repeatedly take two temporally adjacent marble sets, and merge them into a single set of marbles with trajectories of doubled length.
Each iteration of the merging involves a short optimization, where we use rendering losses, tracking losses, and geometric regularizations, to guide the marble sets into correspondence.
At inference, we use the learned Gaussian trajectories to render into a timestep.
In this section, we provide a detailed outline of this optimization strategy. 

\subsection{Dynamic Gaussian Marbles}

\subsubsection{Definition}
Following~\citet{kerbl20233Dgaussians}, our scene representation is a set of Gaussians, $\mathcal{G}$. Different from the original formulation, our Gaussians are \textit{isotropic}:  each Gaussian's orientation is the identity matrix (\textit{i.e.} $R = \mathbf{I}$), and the scale can be written as a scalar value (\textit{i.e.} $s \in \mathbb{R}^1$). To emphasize their spherical shape, we use the name Gaussian \textit{marbles}. We also assign each Gaussian marble to a semantic instance denoted as $y \in \mathbb{N}$ (where instances are provided by an off-the-shelf image segmentation method, described later). Finally, to make each Gaussian marble ``dynamic'', we equip it with a trajectory, represented as a sequence of translations mapping from its initial position $\mu$ to its position at every other timestep. We denote the sequence of translations over a $T$ frame sequence as $\Delta \mathbf{X} \in \mathbb{R}^{T \times 3}$.

\subsubsection{Why Isotropic Marbles?}
While anisotropic Gaussians are far more expressive, we find that the extra degrees of freedom are poorly suited for the underconstrained monocular setting. We refer to \Cref{fig:why-marbles} as a simple illustration of this phenomenon. In this toy experiment, we train anisotropic Gaussians and Gaussan marbles on a single monocular image for 100K iterations. As observed, anisotropic Gaussians fit the training image in a manner that does not generalize to new views, leading to obvious visual artifacts. In contrast, the simpler marbles generalize to novel views. 

\begin{figure}
    \centering
    \includegraphics[width=0.49\textwidth]{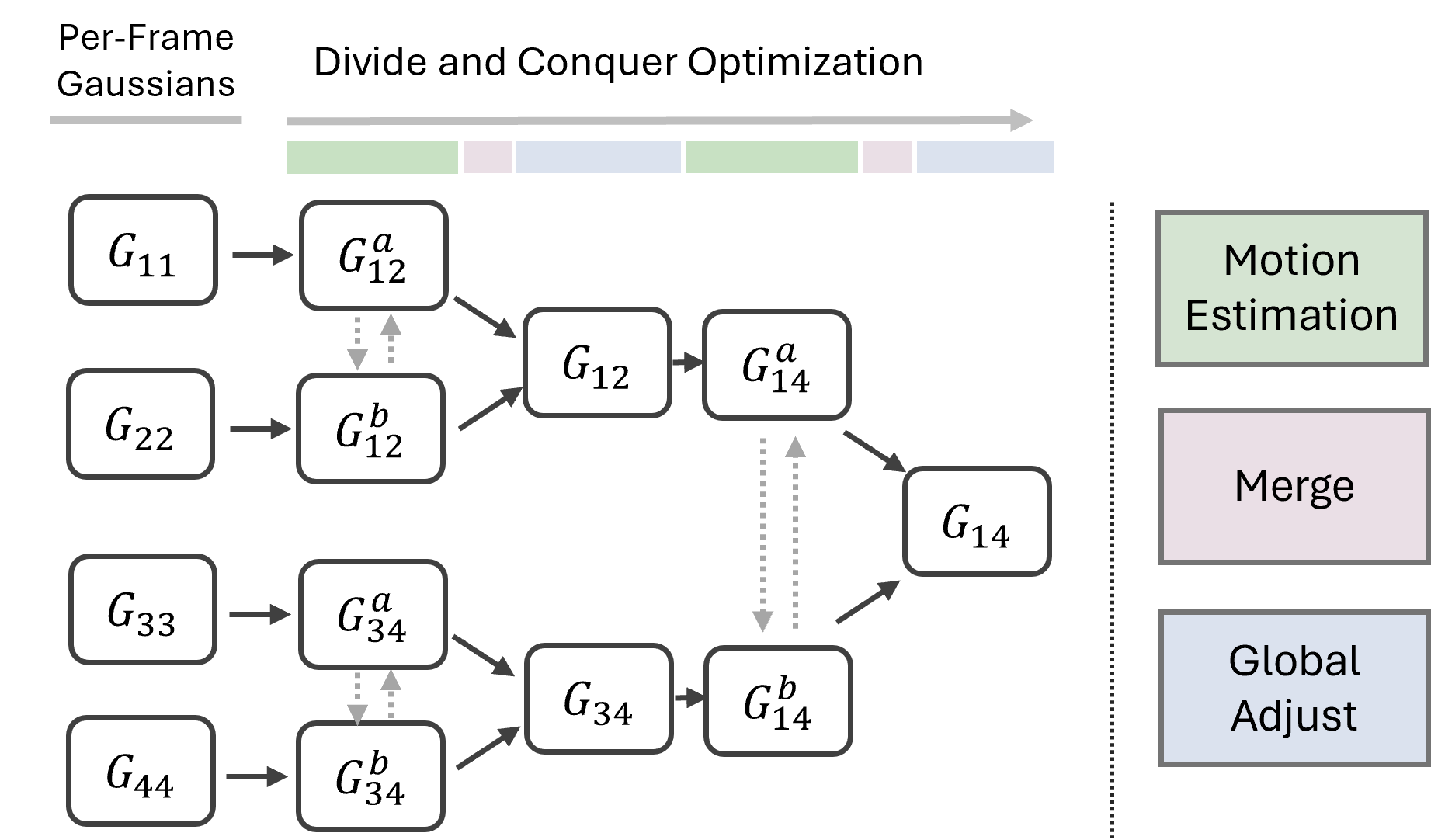}
    \caption{Our divide and conquer learning algorithm iteratively estimates motion between pairs of Gaussian sets, merges the sets, and performs a global adjustment on the Gaussian marbles within the merged sets.}
    \label{fig:hierarchical-motion}
\end{figure}
\subsection{Divide-and-Conquer Motion Estimation}
\subsubsection{Overview}
Rather than attempting to optimize the full video at once, we divide the input video into short subsequences, and run an optimization that iteratively joins subsequences together. 
As shown in \Cref{fig:hierarchical-motion}, this optimization consists of three iterative stages: motion estimation, merging, and global adjustment,
where the length of Gaussian trajectories doubles after each iteration. 
%
%

\subsubsection{Initializing Dynamic Gaussian Marbles}
We initialize a distinct set of Gaussian marbles \textit{per frame}, yielding a sequence of Gaussian sets $[\mathcal{G}_{11}, \mathcal{G}_{22}, ..., \mathcal{G}_{TT}]$. $\mathcal{G}_{ij}$ denotes that each Gaussian trajectory only covers the subsequence of frames $i\to j$; thus, our initial $\mathcal{G}_{ii}$ trivially contain trajectories of length 1.

We achieve this initialization as follows. 
For each frame of the video, we obtain a monocular (or LiDAR) depthmap as well as off-the-shelf temporally-consistent segmentations from the SAM-driven TrackAnything model~\cite{kirillov2023segany,yang2023trackanything}. We then unproject the depth map into a point cloud, and perform outlier removal and downsampling. Then, for each point coordinate $p$, we initialize a Gaussian marble with mean $\mu = p$, color $c$ as the pixel color, instance class $y$ as the segmentation prediction, and we follow the original protocol~\cite{kerbl20233Dgaussians} to initialize scales and opacities. Finally, we initialize the sequence of translations $\Delta \mathbf{X} = [\mathbf{0}]$, \textit{i.e.} as a length-1 sequence of 0 translation. 

\subsubsection{Motion Estimation Phase}
While training on a video with $T$ frames, we have a sequence of Gaussian marble sets:
\[
\mathbf{G} = [\mathcal{G}_{1K},\; \mathcal{G}_{(K+1)(2K)}\;, \mathcal{G}_{(2K+1)(3K)}\;, ..., \;\mathcal{G}_{(cK+1)T}]
\]
with each set of Gaussian marbles covering a length $K$ subsequence. 
To reduce notation and create a simple working example, we will proceed in this section using $K=2$ and $T=8$, giving us 
the sequence $\mathbf{G} = [\mathcal{G}_{12},\; \mathcal{G}_{34},\; \mathcal{G}_{56},\; \mathcal{G}_{78}]$. 

In the motion estimation phase, we begin by forming pairs of adjacent Gaussian marble sets, \textit{i.e.} $[(\mathcal{G}^a_{12},\; \mathcal{G}^b_{34}),\; (\mathcal{G}^a_{56},\; \mathcal{G}^b_{78}]$), where $a$ and $b$ denote whether a set appears earlier or later than its partner. Our goal is to learn a mapping for every Gaussian in $\mathcal{G}^a$ into every frame covered by $\mathcal{G}^b$, and vice versa. To learn these motions, we will render $\mathcal{G}^a$ into frames covered by $\mathcal{G}^b$ and apply the gradient update \textit{only} to the Gaussian trajectories in $\mathcal{G}^a$, and vice versa. 

More concretely, in the case of $\mathcal{G}^a_{12}$, we start by extending the trajectories using a constant-velocity assumption, resulting in an expanded trajectory
\[
\Delta \textbf{X} = [\Delta \textbf{x}_1, \Delta \textbf{x}_2] \; \;
\to \;\; \Delta \textbf{X} = [\Delta \textbf{x}_1, \Delta \textbf{x}_2, \Delta \textbf{x}_3^{\text{init}}]
\]
We then render $\mathcal{G}^a_{12}$ into frame $3$, and compute our optimization objectives (to be described in the next subsection), and backpropagate gradient updates into $\Delta \textbf{x}_3$, \textit{i.e.} the translation into frame 3. We end this optimization after a fixed number of iterations, $\eta$. 
We repeat this for each missing frame in the sequence,  
until we have a trajectory that covers \textit{all} frames in $\mathcal{G}^b$.

\subsubsection{Merging}
The result of motion estimation is that we have two sets of Gaussian marbles which reconstruct the \textit{same} subsequence.
In other words, each pair $(\mathcal{G}^a_{ij}, \mathcal{G}^b_{ij})$ covers the same interval $[i,j]$. Therefore, we can trivially \textit{merge} the pair by taking the union of all the Gaussian marbles, $\mathcal{G}_{ij} = \mathcal{G}^a_{ij} \cup \mathcal{G}^b_{ij}$, yielding a set twice the size of the original sets. 
To avoid excessive computational burden, we drop Gaussians of low opacity and small scale, and additionally perform random downsampling, to keep the set size constant. 

\subsubsection{Global Adjustment Phase}
After merging sets of Gaussian marbles, there is no guarantee that the new resulting set still satisfies our optimization objectives. 
Thus, we jointly optimize \textit{all} Gaussian properties of the newly merged set. Specifically, for the new $\mathcal{G}_{ij}$, we repeatedly randomly sample a frame within $[i,j]$, and render all Gaussians into this frame, and we backpropagate gradient updates to Gaussian colors, scales, opacities, and trajectory translations. We repeat this global adjustment for $\beta$ iterations. 

\subsubsection{Why divide and conquer?}
Our divide and conquer strategy guides the underconstrained optimization problem toward finding solutions which are more realistic. In particular, the motion estimation benefits from the locality and smoothness of adding one additional frame at-a-time, similar to Dynamic 3D Gaussians ~\cite{luiten2023dynamic}, while the global adjustment phase contributes global coherence, similar to 4D Gaussians~\cite{wu20234dgaussians}. By alternating between the two phases, we aim to get the best of both worlds. 


\begin{table*}[]
\setlength{\tabcolsep}{4pt}
    \centering
    \small
    \begin{tabular}{l| c c c c c c c | c}
    \toprule
    \multicolumn{3}{l}{\textbf{\textit{DyCheck iPhone}} - \textit{Without Camera Pose}} & & & & & \multicolumn{2}{l}{\textit{Reported: mPSNR} $\uparrow$ / \textit{LPIPs} $\downarrow$}
    \\
          & Apple & Block & Spin & Paper Windmill & Space-Out & Teddy & Wheel & Mean \\
         Dyn. Gaussians & 7.96 / 0.775 & 7.13 / 0.737 & 9.15 / 0.635 &  6.732 / 0.736 & 7.42 / 0.698 & 7.75 /0.709 & 7.03 / 0.641 & 7.60 / 0.704 \\
         4D Gaussians & 14.44 / 0.716 & 12.30 / 0.706 & 12.77 / 0.697 &  14.46 / 0.790 & 14.93 / 0.640 & 11.86 / 0.729 & 10.99 / 0.803 & 13.11 / 0.726 \\
        Marbles (ours) & \textbf{16.50} / \textbf{0.499} & \textbf{16.11} / \textbf{0.363} & \textbf{17.51} / \textbf{0.424} & \textbf{16.19} / \textbf{0.454} & \textbf{15.97} / \textbf{0.437} & \textbf{13.68} / \textbf{0.443} & \textbf{14.58} / \textbf{0.389} & \textbf{15.79} / \textbf{0.428} \\

        \multicolumn{5}{l}{\textit{ \; \; \; \; \;\;\;\;\;\;\;\;\;\;\;\;\;\;\;\;\;\;- With Camera Pose}}\\
        Dyn. Gaussians & 7.65 / 0.766 & 7.55 / 0.684 & 8.08 / 0.651 & 6.24 / 0.729 & 6.79 / 0.733 & 7.41 / 0.690 & 7.28 / 0.593 & 7.29 / 0.692 \\
        4D Gaussians & 15.41 / \textbf{0.450} & 11.28 / 0.633 & 14.42 / \textbf{0.339} & 15.60 / \textbf{0.297} & 14.60 / \textbf{0.372} & 12.36 / 0.466 & 11.79 / 0.436 & 13.64 / 0.428 \\
        Marbles (ours) & \textbf{17.70} / 0.492 & \textbf{17.42} / \textbf{0.384} & \textbf{18.88} / 0.428 & \textbf{17.04} / 0.394 & \textbf{15.94} / 0.435 & \textbf{13.95} / \textbf{0.442} & \textbf{16.14} / \textbf{0.351} & \textbf{16.72} / \textbf{0.418} \\\midrule
        \multicolumn{5}{l}{\textbf{\textit{Nvidia Dynamic Scenes}}}\\
         & Balloon1 & Balloon2 & Jumping & Playground & Skating & Truck & Umbrella & Mean \\
        Dyn. Gaussians & 8.68 / 0.660 & 13.70 / 0.375 & 11.11 / 0.592 & 11.91 / 0.424 & 13.32 / 0.449 & 15.58 / 0.377 & 10.20 / 0.743 & 12.07 / 0.517\\
        4D Gaussians & 14.11 / 0.404 & 18.56 / 0.239 & 17.32 / 0.326 & 13.51 / 0.341 & 19.41 / 0.218 & 21.25 / 0.172 & 19.00 / 0.346 & 17.59 / 0.292  \\
        Marbles (ours) & \textbf{23.65} / \textbf{0.072} & \textbf{21.60} / \textbf{0.142} & \textbf{19.61} / \textbf{0.180} & \textbf{16.21} / \textbf{0.235} & \textbf{24.24} / \textbf{0.091} & \textbf{27.18} / \textbf{0.060} & \textbf{23.76} / \textbf{0.123} & \textbf{22.32} / \textbf{0.129}\\\midrule


    \bottomrule

    \end{tabular}
    \caption{We report PSNR and LPIPs metrics of Gaussian Marbles and Gaussian baselines on the DyCheck iPhone dataset with pose, the iPhone dataset without camera pose, and the Nvidia Dynamic Scenes dataset. Overall, Gaussian Marbles significantly outperforms previous Gaussian baselines.}
    \label{tab:gaussian-eval}
\end{table*}

\subsection{Losses}
At each optimization step of our divide-and-conquer algorithm, we employ of a variety of loss terms to help drive the Gaussians towards a realistic factorization of scene geometry and motion.

\subsubsection{Tracking Loss}
Building off of recent advances in point tracking ~\cite{harley2022pips, karaev2023cotracker}, we regularize the Gaussian marble trajectories to agree with off-the-shelf 2D point tracks. When optimizing Gaussian trajectory translations into a timestep $j$, we use CoTracker~\cite{karaev2023cotracker} to estimate a set of 
point tracks, $P$, that span the subsequence of frames $[j-w, j+w]$ (with $w=12$ in practice). 
Then, we randomly sample a source frame $i \in [j-w, j+w]$, and regularize the projected Gaussian trajectories to match each tracked point's location at frames $i$ and $j$.

Concretely, we sample 3D Gaussian trajectory positions at frames $i$ and $j$ and project the Gaussians into the image plane, computing the Gaussian 2D means, depths, and 2D covariances.
%
Then, for each tracked point $p_{i\to j}$ in frames $i$ and $j$, 
we find the $K=32$ nearest Gaussians in 2D, and compute a loss that discourages these Gaussians from changing their distance to the tracked point:
\begin{equation}
    \mathcal{L}_{\text{track}} = \sum_{p \in P} \sum_{g \in \mathcal{N}(p_i)} \alpha'_i \; \big\Vert \; D_i ||\mu'_i - p_i||_2 - D_j ||\mu'_j - p_j||_2 \; \big\Vert,
\end{equation}
where  $\mu'_i$ and $D_i$ are the projected location and depth of a Gaussian center $\mu_i$, $\alpha'_i$ is the Gaussian's opacity contribution to $p_{i}$, $P$ is the set of point tracks, and $\mathcal{N}(p)$ is the set of Gaussians that neighbor a pixel $p$.

\subsubsection{Rendering Losses}
At each training iteration, we render an image, disparity map, and segmentation map. For each, we compute a standard L1 loss with the ground truth image, the initial disparity estimation, and the off-the-shelf instance segmentation, as well as a standard LPIPs loss with the ground truth image.

\subsubsection{Geometry Losses}
\paragraph{Isometry Loss}
Following previous works ~\cite{prokudin2023dynamicpointfields, luiten2023dynamic}, we regularize our Gaussian marbles to follow locally rigid motion. In particular, we penalize Gaussians for moving in a manner that breaks isometric deformation on local neighborhoods. Specifically, when rendering into frame $j$, we select a neighboring source timestep $i \in [j-1, j+1]$ and compute a local neighborhood isometry loss as follows:
\begin{equation}
    \mathcal{L}_{\text{iso-local}} = \sum_{g^a \in \mathcal{G}} \sum_{g^b \in \mathcal{N}(g^a)} \big| \Vert \mu^a_{i} - \mu^b_{i} \Vert - \Vert \mu^a_{j} - \mu^b_j \Vert  \big|
\end{equation}
where $\mu^a_i$ and $\mu^b_{i}$ are the means of the Gaussian marbles $g^a$ and $g^b$ at timestep $i$. 

In addition to local isometry, we incorporate an instance isometry loss that guides each unique semantic instance to move in a nearly-isometric manner. That is, when rendering into frame $j$, we select a random frame $i$ and compute the instance isometry loss as follows:
\begin{equation}
    \mathcal{L}_{\text{iso-instance}} = \sum_{g^a \in \mathcal{G}} \sum_{g^b \in Y(g^a)} \big| \Vert \mu^a_{i} - \mu^b_{i} \Vert - \Vert \mu^a_{j} - \mu^b_j \Vert  \big|
\end{equation}
where $Y(g)$ denotes Gaussians with the same semantic label. Our final isometry loss is a combination of the local and instance losses.

\paragraph{3D Alignment Loss}
When merging two distinct sets of Gaussian marbles, it is important that the two sets not only align in the projected image plane, but also in 3D space. Without guiding the optimization towards 3D alignment, it is possible for the resulting merge to be ``cloudy'' in 3D (and in novel views), even if the training-view 2D projection is sharp. This is especially the case when off-the-shelf depth estimation is inconsistent across time.


To achieve 3D alignment, we use a Chamfer loss that pushes Gaussians towards one another. However, because a pair of mergeable Gaussian sets $\mathcal{G}^a$ and $\mathcal{G}^b$ may observe different parts of the scene, we \textit{cannot} directly compute a Chamfer loss between the two. Instead, we divide each set $\mathcal{G}^a$ and $\mathcal{G}^b$ into subsets that contain Gaussians \textit{originating} from a single frame. For example, we would divide the sets $\mathcal{G}^a_{12}$ and $\mathcal{G}^b_{34}$ into subsets $[\mathcal{G}^a_{1}, \mathcal{G}^a_{2}, \mathcal{G}^b_{3}, \mathcal{G}^b_{4}$], where $\mathcal{G}^a_{1}$ contains Gaussians initialized from frame 1. Then, we randomly shuffle the list of subsets and group the first 25\% of shuffled subsets into a set $\mathcal{G}^1$, and the next 25\% into a set $\mathcal{G}^2$. Finally, we compute a 2-way Chamfer distance on $\mathcal{G}^1$ and $\mathcal{G}^2$:
\begin{equation}
    \mathcal{L}_{\text{chamfer}} = \sum_{g^1 \in G^1} \min_{g^2 \in G^2} \big\Vert \mu^1 - \mu^2 \big\Vert_2 +  \sum_{g^2 \in G^2} \min_{g^1 \in G^1} \big\Vert \mu^1 - \mu^2 \big\Vert_2
\end{equation}

This introduces randomness into which parts of the scene are observed in $\mathcal{G}^1$ and $\mathcal{G}^2$, which we find pushes Gaussians closer together without overfitting to differences in observed scene content.



\section{Experiments} \label{sec:experiments}
\begin{table}[]
\setlength{\tabcolsep}{2.5pt}
    \centering
    \small
    \begin{tabular}{l || c c c c c c c | c}
    \toprule
         & \multicolumn{6}{l}{\textit{PCK-T @0.05\%} $\uparrow$}\\
         Method & Apple & Block & Paper & Space & Spin & Teddy & Wheel & Mean \\
    \hline
         Nerfies & 0.318 & 0.216 & 0.107 & 0.859 & 0.115 & 0.775 & 0.408 & 0.400 \\
         HyperNeRf & 0.599 & 0.274 & 0.113 & 0.812 & 0.177 & 0.801 & 0.394 & 0.453 \\
         Dyn. Gauss. & 0.075 & 0.047 & 0.056 & 0.107 & 0.065 & 0.167 & 0.039 & 0.079 \\
         4D Gauss. & 0.000 & 0.000 & 0.000 & 0.229 & 0.033 & 0.133 & 0.076 & 0.073 \\
         \hline
         CoTracker & 0.654 & \textbf{0.907} & 0.926 & \textbf{0.992} & 0.557 & \textbf{0.885} & \textbf{0.703} & \underline{0.803} \\
         \hline
         Ours & \textbf{0.846} & \textbf{0.907} & \textbf{1.000} & \underline{0.947} & \textbf{0.672} & \underline{0.865} & \underline{0.405} & \textbf{0.806} \\

    \bottomrule

    \end{tabular}
    \caption{We report the tracking metric PCK-T @5\% for Gaussian Marbles and baselines on the iPhone dataset in the setting with camera pose.
    }
    \label{tab:tracking-eval}
\end{table}

\subsection{Datasets}
We evaluate our method and a set of competitive baselines on the standard Nvidia Dynamic Scenes ~\cite{Yoon2020nvidiascenes} and DyCheck iPhone~\cite{gao2022dycheck} datasets. However, each of these popular datasets contains multi-view information. Thus, as we will discuss, we modify the training and evaluation protocol to emulate a more monocular setting. Additionally, we evaluate on four scenes from the recent Total-Recon dataset ~\cite{song2023totalrecon}. Finally, our project webpage provides additional qualitative evaluation on videos from the the Davis dataset~\cite{Perazzi2016}, the YouTube-VOS dataset~\cite{Xu2018YouTubeVOSAL}, and on real-world videos.  

\subsubsection{Nvidia Dynamic Scenes Dataset}
The Nvidia Dynamic Scenes dataset~\cite{Yoon2020nvidiascenes} consists of seven videos, each between 90 and 200 frames, and is captured with a rig consisting of 12 calibrated cameras. We evaluate on seven captures - Balloon1, Balloon2, Jumping, Playground, Skating, Truck, and Umbrella. Importantly, previously benchmarked evaluations in the Nvidia dataset sample a \textit{different} training camera at each timestep, resulting in a ``monocular teleporting camera''~\cite{gao2022dycheck}. We consider this setting unrealistic, and hence we instead use the video stream from a single camera, specifically camera 4 for training. We use video streams from cameras 3, 5, and 6 for evaluation. 

\subsubsection{DyCheck iPhone Dataset}
The DyCheck iPhone dataset~\cite{gao2022dycheck} consists of seven casually-captured iPhone videos, each with up to two novel-view and time-synchronized validation videos. We evaluate on all scenes: Apple, Block, Paper Windmill, Space Out, Spin, Teddy, and Wheel. Unlike the Nvidia Dynamic Scenes dataset, the iPhone dataset is truly monocular, \textit{i.e.} only permitting models to train on a single camera stream. However, the training camera follows a purposeful 3D trajectory that circumnavigates the scene, gathering multiview information. While a valid monocular setting, the camera's calculated motion is not representative of casually-captured videos (e.g., as might be found on YouTube). Thus, we evaluate in two settings. First, we follow the official benchmark and use the video stream and camera motion provided. Second, we remove camera poses, offloading the camera motion into the learned 4D scene representation's dynamics. We find this setting interesting because it simulates additional dynamic content, where previously ``static" regions of the scene now have rigid dynamics equal to the inverse camera motion, which must be solved by the scene representation itself. 

\subsubsection{Total-Recon Dataset}
The Total-Recon dataset ~\cite{song2023totalrecon} consists of long indoor video captures of dogs, cats, and humans, captured with two time-synchronized and calibrated iPads equipped with LiDAR. We follow the official protocol and data splits, and we evaluate novel view synthesis on the scenes cat1-stereo000, cat2-stereo000, cat3-stereo000, and human1-stereo000. 

\subsection{Training and Inference Details}
\begin{table}[]
\setlength{\tabcolsep}{2pt}
    \centering
    \small
    \begin{tabular}{l | c c c | c}
    \toprule
         Method & iPhone \textit{($+$ pose)} & iPhone \textit{($-$ pose)} & Nvidia & Mean \\
    \midrule
         \hline
         Nerfies & 16.45 / 0.339 & 14.60 / 0.483 & 21.40 / 0.190 & 17.48 / 0.337\\
         HyperNeRf & 16.81 / \textbf{0.332} & 14.97 / 0.474 & 21.73 / 0.167 & \underline{17.83} / \textbf{0.324}\\
         T-NeRF & \textbf{16.96} / 0.379 & 14.54 / 0.574 & 21.40 / 0.171 & 17.63 / 0.375 \\
         \hline
         Ours & 16.72 / 0.418 & \textbf{15.79} / \textbf{0.428} & \textbf{22.32} / \textbf{0.129} & \textbf{18.28} / \underline{0.325}\\
    \bottomrule
    \end{tabular}
    \caption{We report mPSNR$\uparrow$ / LPIPs $\downarrow$. Gaussian Marbles is on-par with NeRF baselines for the task of novel view synthesis.}
    \label{tab:nerf-eval}
\end{table}

\subsubsection{Implementation Details}
On the DyCheck iPhone dataset, we initialize each frame with 220,000 Gaussians when camera poses are provided and 180,000 Gaussians when they are not, and we downsample back to these quantities after each merge. We take $\eta=80$ optimization steps per frame during motion estimation and $\beta=32$ steps per frame during global adjustment. We stop our divide-and-conquer curriculum after learning subsequences of length 8 on the foreground, length 32 on the background when camera pose is held out, and length 512 on the background when camera pose is provided. When camera poses are provided, we still learn a dynamic background to account for errors in the provided poses.


On the Nvidia Dynamic Scenes dataset, we initialize each frame with 120,000 Gaussians and upsample to 240,000 Gaussians during the last stage of global adjustment. We take $\eta=128$ optimization steps during motion estimation and $\beta=48$ steps during global adjustment, and we stop our learning at subsequences of length 32 on both the foreground and background.
On the Total-Recon dataset, we initialize each frame with 120,000 Gaussians, set $\eta=80$ and $\beta=32$, and stop our learning at subsequences of length 8 on the foreground and length 32 on the background.

Please refer to the Appendix for additional implementation details, including information on loss weights, downsampling, depth estimation, real-world videos, and more.


\subsubsection{Runtime and Memory}
The runtime and memory requirements of Gaussian Marbles increase with the length of the input video. On the DyCheck iPhone dataset, on a single NVIDIA A5000 GPU, the training curriculum takes 5 hours and peaks at 13GB of memory on the shortest 277 frame ``paper windmill" video, and 9 hours and peaks at 22GB of GPU memory on the longest 475 frame ``apple" video. On the Nvidia dataset, on a single NVIDIA V100 GPU, the training takes 3.5 hours and peaks at 6GB of memory on the shortest 95 frame ``Jumping" video, and 8 hours and peaks at 10GB of memory on the longest 203 frame ``Truck" video. On both datasets, our inference runs at over 200Hz on an NVIDIA RTX-3090 GPU.

\subsection{Dynamic Novel View Synthesis with Gaussians}
We evaluate Gaussian Marbles against recent methods Dynamic 3D Gaussians ~\cite{luiten2023dynamic} and 4D Gaussians ~\cite{wu20234dgaussians}. We report the standard metrics mPSNR and LPIPs on novel view synthesis in ~\Cref{tab:gaussian-eval}. We see that Gaussian Marbles significantly outperforms both Gaussian baselines on average across both datasets. In particular, Gaussian Marbles significantly improves over baselines in settings with less multi-view information, \textit{i.e.} the iPhone evaluation without camera pose and the Nvidia Dynamic Scenes evaluation. 

\begin{figure*}
    \centering
    \includegraphics[width=0.8\textwidth]{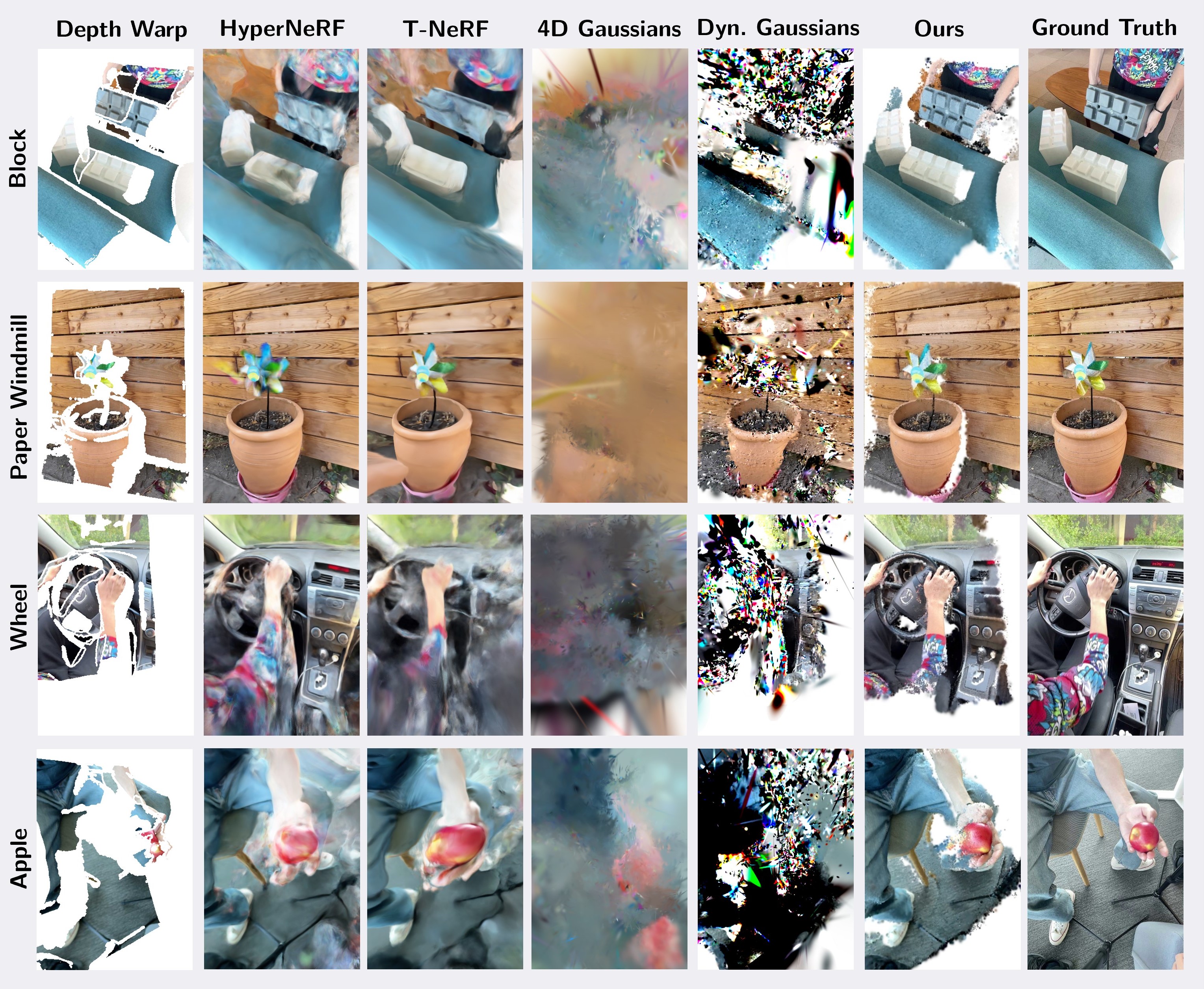}
    \caption{We visualize novel view synthesis Gaussian Marbles and baselines on various scenes of the DyCheck iPhone dataset (setting without camera pose).}
    \label{fig:dycheck-nopose-qual}
\end{figure*}

We visualize the results of Gaussian Marbles and the baselines on the iPhone dataset (without camera pose) in ~\Cref{fig:teaser} and ~\Cref{fig:dycheck-nopose-qual}. As shown, the existing Gaussian baselines exhibit poor novel view synthesis in this monocular setting, further emphasizing their need for strong multi-view supervision. In particular, 4D Gaussians converges to a local minima, averaging the static information over all frames instead of correctly learning motion. On the other hand, the Gaussians in Dynamic Gaussians immediately diverge from the scene geometry, overfitting to the training view in a manner that does not correctly render into novel views. ~\Cref{fig:dycheck-nopose-qual} also offers a qualitative comparison with depth warping ~\cite{Niklaus_CVPR_2020}, showing that Gaussian Marbles aggregates and renders a greater portion of scene's content compared to single-frame warping.

In ~\Cref{fig:tiger-qual}, we qualitatively compare Gaussian Marbles and 4D Gaussians on a real-world video of a tiger. While 4D Gaussians struggles to reconstruct the training sequence, Gaussian Marbles achieves high-quality reconstruction and novel-view synthesis.

\begin{table}[]
\setlength{\tabcolsep}{2pt}
    \centering
    \small
    \begin{tabular}{l | l | c c c c | c}
    \toprule
         Method & Domain & \;Cat1\; & \;Cat2\; & \;Cat3\; & Human1 & \;Mean\; \\
    \midrule
         \hline
         HyperNeRF & Any Motion & 0.532 & 0.330 & 0.514 & 0.501 & 0.469\\
         D$^2$NeRF & Any Motion & 0.685 & 0.561 & 0.730 & 0.585 & 0.640\\
         Ours & Any Motion & \underline{0.446} & \underline{0.284} & \underline{0.386} & \underline{0.388} & \underline{0.376}\\
         \hline
         Total-Recon & Human + Quadruped & \textbf{0.382} & \textbf{0.237} & \textbf{0.261} & \textbf{0.213} & \textbf{0.273} \\

    \bottomrule
    \end{tabular}
    \caption{We report LPIPs $\downarrow$ on the Total-Recon dataset. Gaussian Marbles outperforms 4D NeRF baselines, and begins to close the gap with domain-specific Total-Recon.}
    \label{tab:totalrecon-eval}
\end{table}

\subsection{Tracking and Editing with Gaussians}

In addition to novel view synthesis, Gaussian trajectories are well suited for dense point tracking.  ~\Cref{tab:tracking-eval} presents tracking accuracy on the Dycheck iPhone dataset in the setting with camera poses provided -- we follow the official DyCheck evaluation and report the percentage of correctly tracked keypoints (PCK-T) at a 5\% interval.

As reported, Gaussian Marbles significantly outperforms other NeRF and Gaussian-based methods in tracking. Moreover, Gaussian Marbles shows a slight improvement over off-the-shelf CoTracker predictions. Notably, the scenes "Space Out" and "Wheel", where tracking performance lags behind CoTracker, have been reported to have inaccurate depth estimation ~\cite{som2024}. This suggests that Gaussian Marbles can incorporate 3D information to improve tracking, but fails to do so when 3D depth is unreliable.

\begin{figure*}
    \centering
    \includegraphics[width=0.6\textwidth]{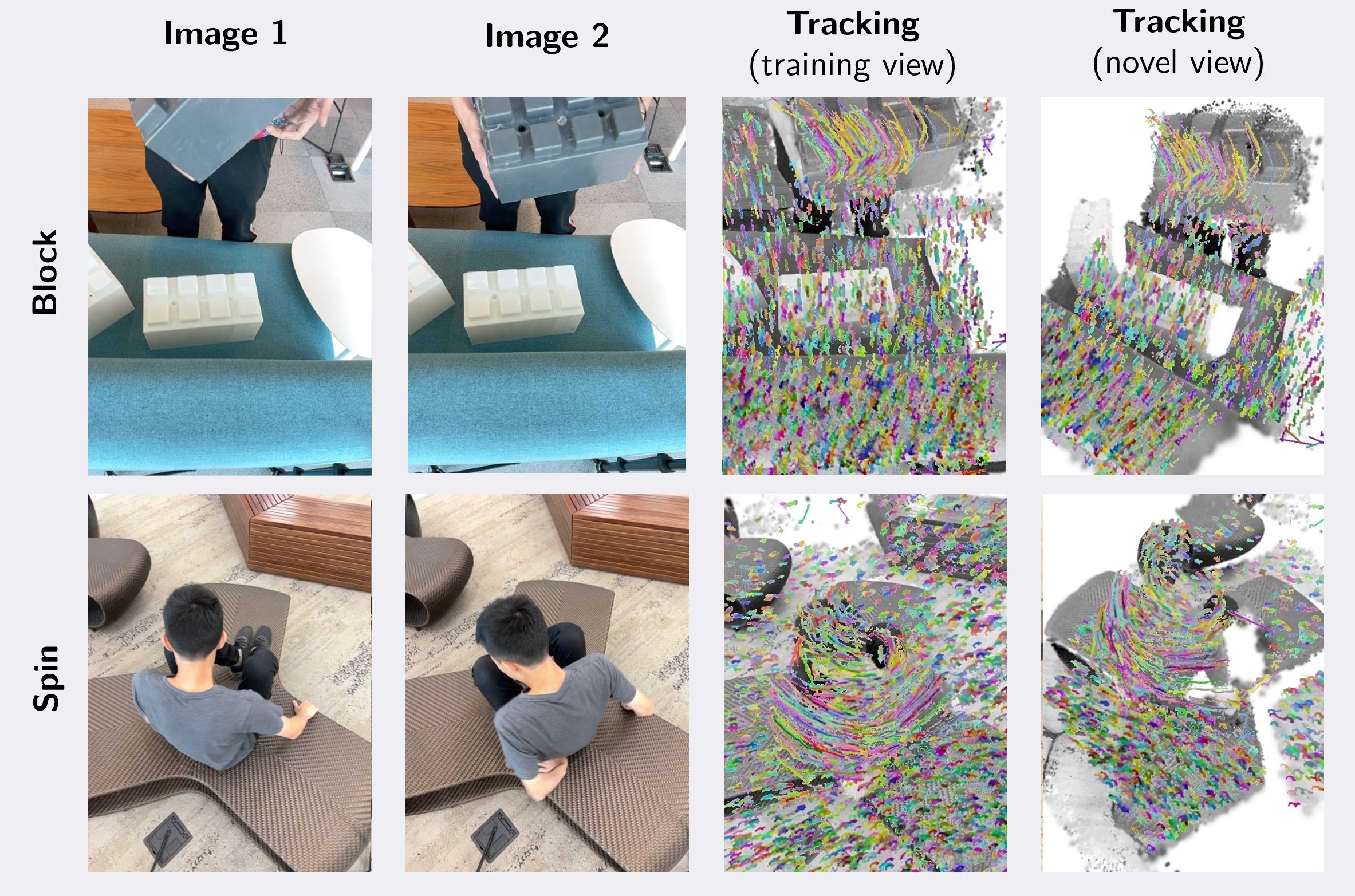}
    \caption{We visualize dense point tracking of Gaussian Marbles on two scenes from the DyCheck IPhone dataset (in the setting where camera pose is withheld).}
    \label{fig:tracking-qual}
\end{figure*}

~\Cref{fig:tracking-qual} visualizes dense Gaussian Marbles point tracks for training and novel views in the setting where camera pose is witheld -- Gaussian Marbles successfully tracks the dense scene geometry of both the foreground and background.
%
In ~\Cref{fig:tiger-qual}, we show that tracking allows the editing videos in a temporally consistent manner. We color the tiger blue in the first frame and propagate the edit throughout the entire video. This emphasizes that dynamic Gaussians are a good choice of scene representation for editing applications.

\subsection{Dynamic Novel View Synthesis with NeRF}
\begin{figure*}[h!]
    \centering
    \includegraphics[width=\textwidth]{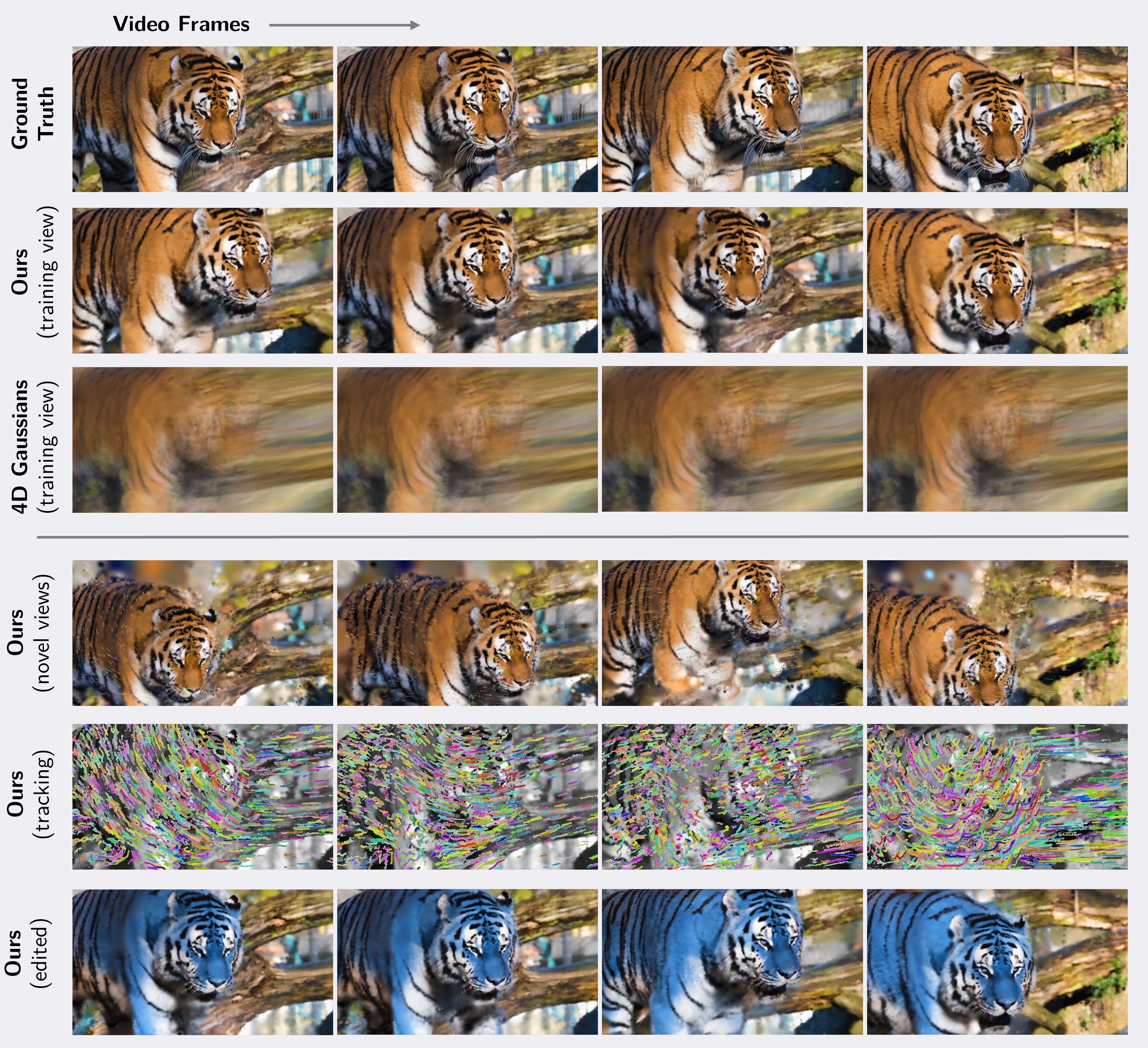}
    \caption{Gaussian Marbles reconstructs a sequence in a manner that tracks and aggregates 3D content, allowing novel-view synthesis with downstream edits.}
    \label{fig:tiger-qual}
\end{figure*}

In ~\Cref{tab:nerf-eval}, we compare Gaussian Marbles with competitive NeRF baselines Nerfies~\cite{park2021nerfies}, HyperNeRF~\cite{park2021hypernerf}, and T-NeRF~\cite{gao2022dycheck} on the DyCheck iPhone and Nvidia datasets. Overall, Gaussian Marbles is on-par with NeRF baselines. Gaussian Marbles does very 
well on the iPhone evaluation without camera pose, and we speculate that the volumetric NeRF approaches struggle in the absence of a static background (\textit{i.e.} when the entire scene moves), while Gaussians can better handle the more expansive dynamic region. 
Furthermore, we again emphasize that Gaussian Marbles exhibits significantly faster rendering (see ~\Cref{fig:teaser}), better tracking (see ~\Cref{tab:tracking-eval}), and more editability (see ~\Cref{fig:tiger-qual}) than the NeRFs.  

We also compare with NeRF baselines on the Total-Recon dataset in ~\Cref{tab:totalrecon-eval}. On this more challenging dataset, Gaussian Marbles significantly outperforms general-purpose dynamic NeRF methods HyperNeRF and D$^2$NeRF~\cite{Wu2022D2NeRFSD}. However, there is still a gap between Gaussian Marbles and the domain-specific Total-Recon, which uses human and quadruped priors to achieve better reconstruction. 

\subsection{Ablations}
In ~\Cref{tab:ablation}, we ablate various parts of Gaussian Marbles and report the outcomes on three scenes from the Nvidia dataset. As shown, each of our design choices helps in achieving high quality novel view synthesis. In particular, the table suggests that both our motion estimation and global adjustment phases are crucial toward achieving high quality scene reconstruction.

\begin{table}[]
\setlength{\tabcolsep}{1.0pt}
    \centering
    \small
    \begin{tabular}{l | c c c | c}
    \toprule
         & \multicolumn{3}{l}{\textit{mPSNR} $\uparrow$ / \textit{mLPIPS} $\downarrow$}\\
         Method & Jumping & Skating & Truck & Mean \\
    \midrule
         No Segmentation &  19.51 / 0.188 & 24.27 / 0.098 &  26.88 / 0.072 & 23.55 / 0.119\\
         No Tracking &  19.50 / 0.198 & 24.21 / 0.093 &  26.98 / 0.065& 23.56 / 0.119 \\
         No Isometry & \textbf{19.61} / 0.186 & \textbf{24.38} / 0.097 & 27.01 / 0.062 & 23.67 / 0.115 \\ \hline
         No Motion Estimation & 19.30 / 0.222 &  23.89 / 0.115 & 26.60 / 0.074 & 19.93 / 0.136 \\
         No Global Adjustment & 19.87 / 0.271 & 24.00 / 0.187 & 25.73 / 0.211 & 19.87 / 0.223 \\\hline
         DGMarbles & \textbf{19.61} / \textbf{0.180} & 24.24 / \textbf{0.091} & \textbf{27.18} / \textbf{0.060} & \textbf{23.68} / \textbf{0.110} \\

    \bottomrule

    \end{tabular}
    \caption{We ablate various components of Gaussian Marbles, showing that each component is important to achieve high quality novel view synthesis.}
    \label{tab:ablation}
\end{table}

\section{Limitations and Conclusion}
We present Gaussian Marbles, an attempt to bring dynamic Gaussians to the challenging setting of casual monocular video captures. Gaussian Marbles introduces isotropic Gaussian ``marbles'', a divide-and-conquer learning strategy, and various 2D priors. Taken together, Gaussian Marbles performs novel view synthesis that is significantly better than previous Gaussian methods. Furthermore, Gaussian Marbles is well-suited for tracking and editing, and significantly outperforms previous reconstruction methods on tracking accuracy. Nevertheless, while Gaussian Marbles presents significant improvements, it has many limitations and does not comprehensively solve the extremely challenging problem of open-world dynamic and monocular novel view synthesis. Since Gaussian Marbles relies on 2D image priors, errors in the 2D predictions such as poor depth estimation, poor segmentation, or poor tracking can lead the optimization into suboptimal results. Similarly, our geometric priors may guide optimization incorrectly in scenes with rapid and non-rigid motion -- a setting where further progress in 3D priors and visual tracking will be vital. 
We hope Gaussian Marbles provides a significant step forward in bringing Gaussian representations to the challenging setting of general monocular novel-view synthesis.

\section*{Acknowledgments} 
This work was supported by grants from the Army Research Laboratory, the Office of Naval Research, and a Vannevar Bush Faculty Fellowship. Additionally, this material is based on work that is partially funded by an unrestricted gift from Google.

\bibliographystyle{ACM-Reference-Format}
\bibliography{bibliography}

\newpage


\twocolumn
\newpage
\appendix

In this document, we provide additional details regarding our method and experiments.

\section{Additional Method Details}

\subsection{Initializing Gaussian Marbles}
For each frame of the input video, we initialize a set of Gaussians with a point cloud obtained from unprojecting per-frame depth maps. On the DyCheck iPhone dataset and Total-Recon datasets, the depth maps are provided from the iPhone or iPad LiDAR. On the Nvidia Dynamic Scenes Dataset, we estimate a depth map using DepthAnything~\cite{yang2024depthanything}. Then, to ensure that the monocular depth estimates have the correct shift and scale for proper novel view evaluation, we solve for a per-frame shift and scale using a ground truth point cloud estimated by running COLMAP on all 12 calibrated cameras. On our qualitative evaluation on videos from the DAVIS dataset, YouTube-VOS dataset, and the internet, we use DepthAnythingV2~\cite{depth_anything_v2} to estimate metric depth. 

After unprojecting the depth map into a point cloud, we perform outlier removal by calling Open3D's ``remove statistical outliers" with 20 neighbors and a standard deviation of 2.0. Then, we downsample the filtered point cloud to the desired number Gaussians by randomly selecting points with probability inversely proportional to their depth. By downsampling in this manner, we select more points that are closer to the camera and less points that are farther away from the camera.

\subsection{Motion Estimation Phase}
During the motion estimation phase, when rendering a Gaussian set $\mathcal{G}^a$ into a frame covered by Gaussain set $\mathcal{G}^b$, with probability $0.5$, we will also render Gaussians from $\mathcal{G}^b$. However, we do not propagate gradient updates back to $\mathcal{G}^b$ in this setting. During motion estimation, we set the learning rate of learned trajectory translations $\Delta X$ to be $0.002$, although find the optimization is relatively robust to this parameter.

\subsection{Merging}
At each merge, we prune and downsample the merged Gaussian set to maintain a constant number of Gaussians. In pruning, we remove any Gaussians with an opacity below $0.02$ or a scale below $0.002$. The downsampling is a straightforward random downsample. After merging, we qualitatively find it slightly favorable to reduce the scale of each Gaussian to 85\% of its original size, although note that this has a marginal effect.

\subsection{Global Adjustment Phase}
During global adjustment, with probability $0.5$, we randomly dropout 50\% of the Gaussians. We set the Gaussian color learning rate to $0.0025$, opacity learning rate to $0.05$, scaling learning rate to $0.002$, and trajectory translation $\Delta X$ learning rate to $0.0004$ (5 times smaller than in motion estimation).

\begin{figure}
    \centering
    \includegraphics[width=0.45\textwidth]{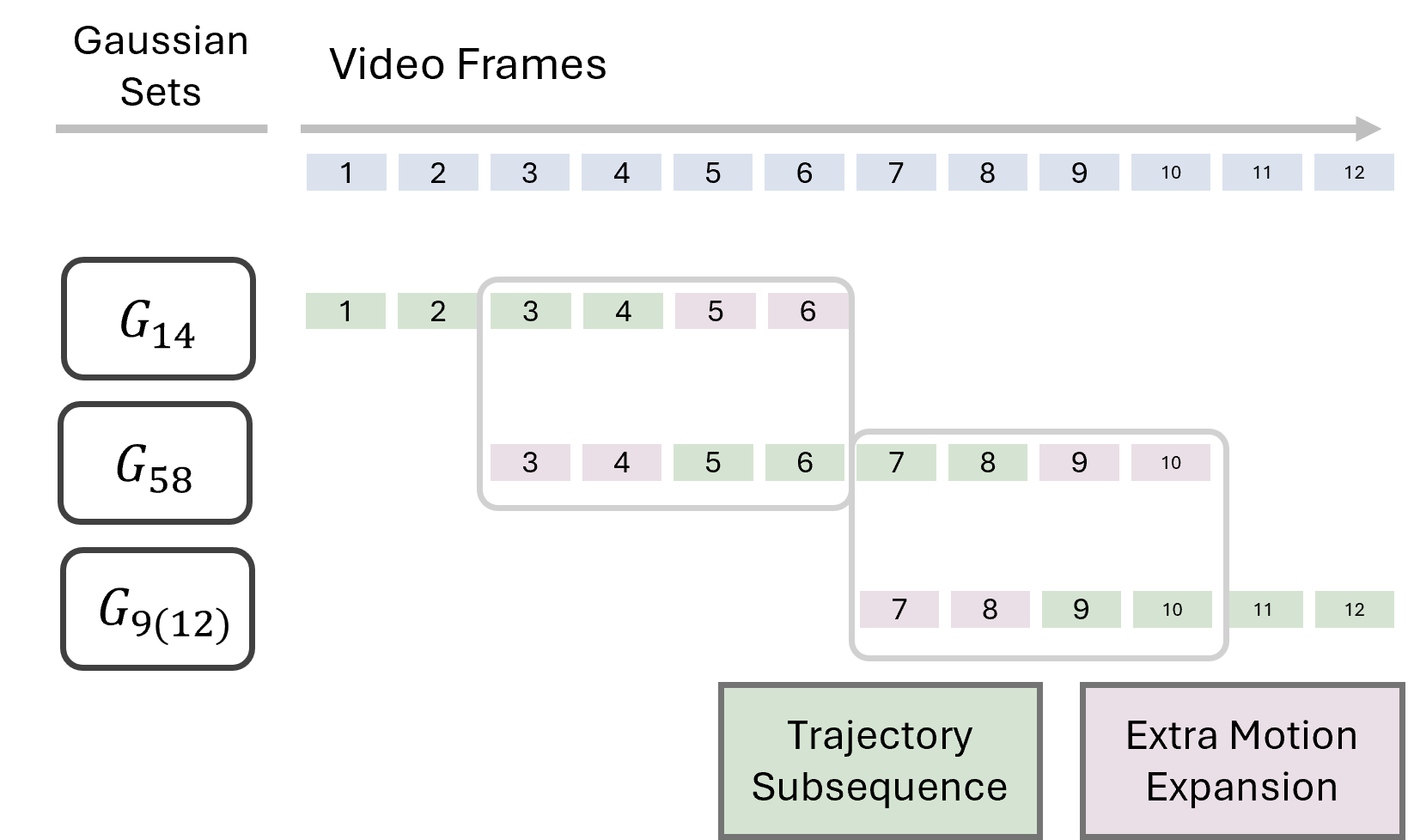}
    \caption{To avoid sudden representational changes at the boundaries of Gaussian trajectory subsequences, we perform an additional phase of motion estimation. After this phase, every frame is covered by two sets of Gaussian trajectories, and we can smoothly update scene content via a sliding window of trajectories.}
    \label{fig:sliding-window-expansion}
\end{figure}

\subsection{Rendering a Sliding Window at Inference}
Our divide-and-conquer learning strategy creates many Gaussian sets, where each set contains trajectories that span a unique and non-overlapping subsequence of the input video. Each of these Gaussian sets can be used for high quality novel view synthesis on frames within the covered subsequence. However, we found that it is qualitatively jarring to suddenly transition between Gaussian sets (\textit{i.e.} at the boundaries between subsequences). 

To address this, we perform an additional phase of motion estimation \textit{after} our divide and conquer.
We illustrate the additional motion estimation in \Cref{fig:sliding-window-expansion}. This additional phase is identical to any other motion estimation phase, except we expand the Gaussian trajectories forward in time \textit{and} backward in time, for a number of frames equal to half the original sequence length. For instance, the set of length-8 Gaussian trajectories $\mathcal{G}_{8(16)}$ would be expanded to length 16 Gaussian trajectories $\mathcal{G}_{4(20)}$. 

After this motion expansion, every frame in the video is covered by exactly \textit{two} sets of Gaussian trajectories (except at the start and end of the video). At inference, we sample Gaussians from \textit{both} sets in a sliding window fashion. That is, we sample Gaussians that were initialized from the $K$ frames \textit{closest} to the current frame. We find this slightly harms single frame rendering quality, but improves the temporal coherence at subsequence boundaries. The reported metrics in the main paper use this sliding window approach - while we could have achieved better metrics without it, we believe the qualitative temporal coherence is more important.

\subsection{Tracking at Test Time}
We wish to track Gaussians through an entire video, not just a subsequence of the video. To achieve this, we ``glue together" trajectories at the boundaries of learned subsequences. As a working example, suppose we have Gaussian sets $\mathcal{G}_{14}$ and $\mathcal{G}_{58}$. To track from frame 1 to frame 8, we forecast Gaussians from $\mathcal{G}_{14}$ into frame 5 using a constant velocity assumption, and then identify corresponding Gaussians in $\mathcal{G}_{58}$ using a nearest neighbor lookup. In practice, we find it best to only search for neighbors within the same segmentation class, and we define neighbor distance as a combination of 3D distance and projected 2D pixel distance.

\subsubsection{Additional Details on DyCheck Evaluation}
Following the official dataset protocol, we train and evaluate with images at 360x480 resolution. For rendering losses, we set the disparity loss weight to 0.1, segmentation loss weight to 0.4, photometric loss weight to 0.1, and LPIPs loss weight to 0.001. We additionally set the tracking loss weight to 0.03, the isometry loss weight to 10.0, and the instance isometry loss to 2.0. For the DyCheck dataset, we do not use our 3D alignment Chamfer loss as we find the LiDAR depth maps are metrically consistent enough. 

\subsubsection{Additional Details on Nvidia Evaluation}
Following previous protocols, we rescale all images to a height of 248 pixels and a width that maintains the original aspect ratio. For rendering losses, we set the disparity loss weight to 0.1, segmentation loss weight to 0.4, photometric loss weight to 0.7, and LPIPs loss weight to 0.015. We additionally set the tracking loss weight to 0.01, the isometry loss weight to 10.0, and the instance isometry loss to 0.0. We do not use our 3D alignment Chamfer loss as we again find the COLMAP-scaled depth maps are sufficiently consistent. 

\subsubsection{Additional Details on Total-Recon Evaluation}
Following official evaluation, we train and evaluate with images at 360x480 resolution. For rendering losses, we set the disparity loss weight to 0.05, segmentation loss weight to 0.4, photometric loss weight to 0.3, and LPIPs loss weight to 0.01. We additionally set the tracking loss weight to 0.01, the isometry loss weight to 5.0, and the instance isometry loss to 0.1. We do not use our 3D alignment Chamfer loss as we find the LiDAR depth maps are metrically consistent.

\subsubsection{Additional Details on Real World Videos}
Please refer to our project website for results on real world videos. On real world videos from the DAVIS dataset, YouTube-VOS dataset, and online, we initialize each frame with 300,000 Gaussians and downsample back to 300,000 after merges. We take $\eta=160$ optimization steps per frame during motion estimation and $\beta=64$ steps per frame during global adjustment. We stop our divide-and-conquer curriculum after learning subsequences of length 32 on the foreground and background. We also experimented with length 128 and found that there is an observable, but not excessive, drop in quality -- therefore, for applications that require longer-term tracking such as edit propagation, we suggest continuing the curriculum up to 128-frame subsequences. 
For rendering losses, we set the disparity loss weight to 0.06, segmentation loss weight to 0.4, photometric loss weight to 0.2, and LPIPs loss weight to 0.002. We additionally set the tracking loss weight to 0.008, the isometry loss weight to 10.0, the instance isometry loss to 0.5, and the 3D alignment Chamfer loss weight to 0.2. While we can handle arbitrary resolution, we generally try to downsize images to be close to standard resolution (640x480).


\bibliographystyle{ACM-Reference-Format}
\bibliography{bibliography}


\end{document}